\documentclass[letterpaper, 10 pt, conference]{ieeeconf}

\IEEEoverridecommandlockouts	

\usepackage{graphicx} 
\usepackage{amsmath,amssymb,amsbsy}  
\usepackage[utf8]{inputenc}
\usepackage{todonotes,cite,booktabs} 
\usepackage[hidelinks]{hyperref} 
\usepackage{siunitx,textcomp} 
\usepackage[caption=false]{subfig}

\let\labelindent\relax
\usepackage[inline]{enumitem} 

\usepackage[nolist]{acronym} 
\newacro{bn}[BN]{Batch Normalization}
\newacro{relu}[ReLU]{Rectified Linear Unit}
\newacro{rmsle}[RMSLE]{Root Mean Squared Logarithmic Error}
\newacro{mse}[MSE]{Mean Squared Error}
\newacro{adam}[Adam]{Adaptive Moment Estimation}
\newacro{dnn}[DNN]{Deep Neural Network}
\newacro{cnn}[CNN]{Convolutional Neural Network}
\newacro{rnn}[RNN]{Recurrent Neural Network}
\newacro{resnet}[ResNet]{Residual Neural Network}

\usepackage{standalone,tikz,pgfplots} 
\pgfplotsset{compat=1.13}

\newcommand{\figref}[1]{\hyperref[#1]{Fig.~\ref*{#1}}}
\newcommand{\tabref}[1]{\hyperref[#1]{Tab.~\ref*{#1}}}
\newcommand{\secref}[1]{\hyperref[#1]{Sect.~\ref*{#1}}}
\newcommand{\algoref}[1]{\hyperref[#1]{Alg.~\ref*{#1}}}
\newcommand{\matr}[1]{\mathbf{#1}}
\newcommand{\unit}[1]{\ensuremath{\, \mathrm{#1}}}
\newcommand{\includeexample}[2]{
	\subfloat[#2]{
		\begin{tabular}{c}
			\includegraphics[width=.19\linewidth]{Figures/#1_photo}\\[-.15em]
			\includegraphics[width=.19\linewidth]{Figures/#1_rviz}
		\end{tabular}\label{fig:examples_#1}
	}}
\def\laserlegend{the green laser is the original input signal, while the red 
one is the output produced in real-time by the proposed algorithm}
\def\bestcolor{(best viewed in color)}
\def\truedist{robot-to-obstacle distance}
\def\ie{\textit{i.e.,}}
\def\eg{\textit{e.g.,}}
\def\etal{\textit{et al.}}
\def\cob{Care-O-bot 4}
\def\true{\checkmark}
\def\false{}

\title{\LARGE \bf
Hallucinating Robots:\\Inferring Obstacle Distances from Partial Laser 
Measurements}

\author{Jens~Lundell\textsuperscript{*}, Francesco~Verdoja\textsuperscript{*} 
and Ville~Kyrki%
\thanks{\textsuperscript{*}These authors contributed equally to this paper.}%
\thanks{This work was supported by Academy of Finland, decision 314180.}%
\thanks{J.~Lundell, F.~Verdoja and V.~Kyrki are with School of Electrical 
Engineering, Aalto University, Finland. \url{name.surname@aalto.fi}}}

\begin{document}

\maketitle
\thispagestyle{empty}
\pagestyle{empty}


\begin{abstract}
Many mobile robots rely on 2D laser scanners for localization,
mapping, and navigation. However, those sensors are unable to
correctly provide distance to obstacles such as glass panels and
tables whose actual occupancy is invisible at the height the sensor is
measuring. In this work, instead of estimating the distance to
obstacles from richer sensor readings such as 3D lasers or RGBD
sensors, we present a method to estimate the distance directly from raw 2D
laser data. To learn a mapping from raw 2D laser distances to obstacle
distances we frame the problem as a learning task and train a neural
network formed as an autoencoder. A novel configuration of network
hyperparameters is proposed for the task at hand and is quantitatively
validated on a test set. Finally, we qualitatively demonstrate in real
time on a \cob{} that the trained network can successfully infer
obstacle distances from partial 2D laser readings.
\end{abstract}


\section{Introduction}
\label{sec:intro}

\begin{figure}
	\includegraphics[width=.49\linewidth]{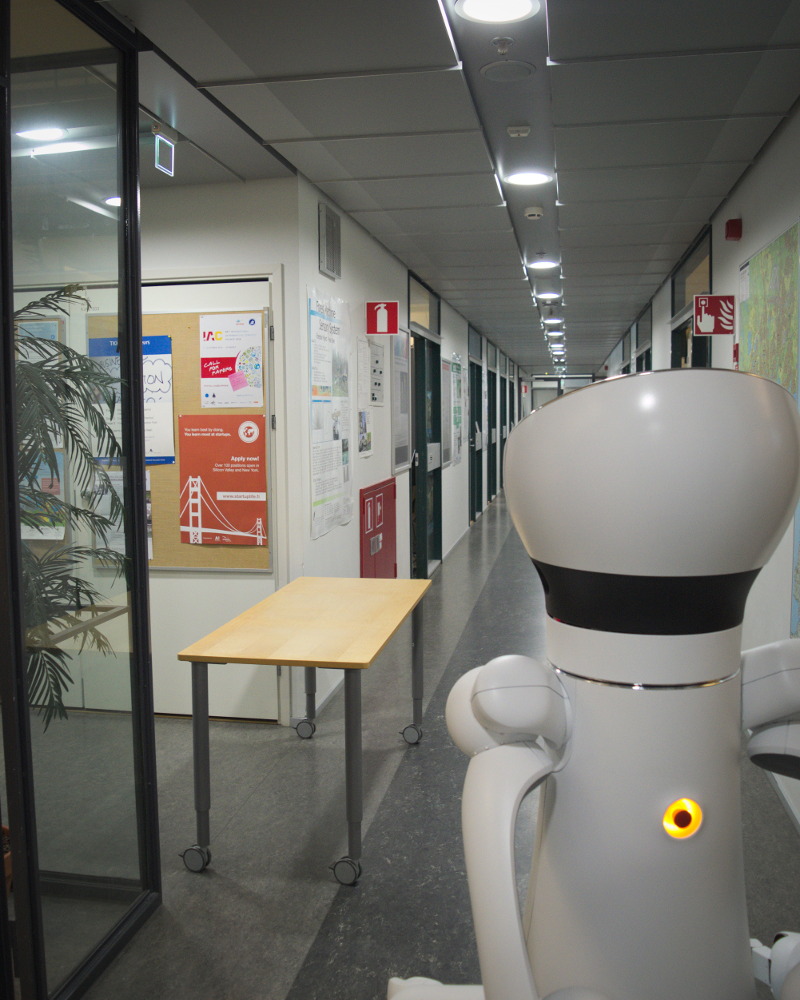}
	\includegraphics[width=.49\linewidth]{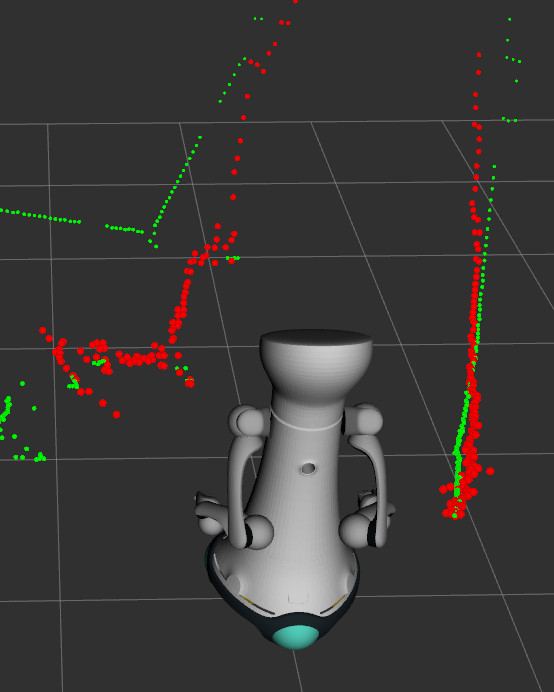}
	\caption{\label{fig:cob}A robot reconstructing the shape of a table by 
		seeing only its legs. In the illustration on the right, \laserlegend{} 
		\bestcolor{}.}
\end{figure}

2D laser scanners are the de facto sensors used by mobile robots for
navigation, mapping, and localization as they provide distance
measurements in large angular fields at fast rates thanks to their
small data dimensionality \cite{baltzakis_fusion_2003}. However, the
amount of knowledge that can be extracted from a 2D laser scan may be
insufficient for some tasks like object detection and obstacle
avoidance. In particular, laser sensors cannot detect glass or infer
the true occupancy of complex objects such as tables. An example of
this problem is presented in \figref{fig:cob}: here a robot relying
only on raw 2D laser data of the scene (\ie{} the green dots in the
figure) would see the legs of the table but not the tabletop itself,
allowing it to plan and execute a trajectory through the table causing
a collision and harming the robot. We refer to the distance along a
certain direction to the closest point of an obstacle that the robot
could collide with as the \truedist{}.

Classical approaches overcome such limitations by utilizing richer sensor 
readings from, for example, RGBD cameras \cite{bonin_font_visual_2008} or 3D 
laser scanners \cite{surmann_autonomous_2003}, or by fusing data between 
different sensors \cite{baltzakis_fusion_2003}. These approaches often suffer 
from a limited field of view and higher computational requirements of the 
sensors they employ.
Instead, we propose a method for inferring the \truedist{} directly from 2D 
laser data. 
This is accomplished by framing the problem as a learning 
task where the input is raw 2D laser distance and the output is the correct 2D 
\truedist{}. In essence, the network \emph{hallucinates} the 
corresponding range data as if the 2D laser scanner could
detect the full shape of the obstacles; hence the name \emph{Hallucinating 
robots}\footnote{Code of the implementation of the proposed 
method, the dataset used in this study, and a video demonstrating the 
algorithm running in real time on a robot can be found on our  
website: 
\url{http://irobotics.aalto.fi/software-and-data/hallucinating-robots}}. 

The main contributions of this work are:
\begin{itemize}
	\item a novel approach to estimate the \truedist{} from raw 2D laser data 
	by training an autoencoder; 
	\item a quantitative study over the impact on the network performance of 
	different hyperparameters, specifically skip connections 
	(\secref{sec:skip}), a novel 	activation function tailored for the domain 
	at hand (\secref{sec:gamma}) and data augmentation with random Gaussian 
	noise (\secref{sec:augmentation});
	\item a method to generate ground truth \truedist{}s by fusing 
	raw 2D laser scans with overlapping depth images (\secref{sec:dataset});
	\item a qualitative evaluation of the proposed technique performed online 
	on the real robot (\secref{sec:qualitative_eval}) showing the method 
	working in complex 	environments such as the one presented in 
	\figref{fig:cob} and in \ang{360} around the robot.
\end{itemize}

To the best of our knowledge, the work presented here is the first one focusing 
on learning distance to otherwise invisible obstacles in 2D laser data. 
The hallucinated laser data is not to replace raw laser data, but instead acts 
as an additional source of information to enrich environment awareness and 
increase robot safety.

\section{Related works}
\label{sec:related}

One of the earliest works on training neural networks on laser data dates back 
to 1998 \cite{burgard_interactive_1998}, where the trained network was used in 
a museum tour-guide robot to modify the initial map with the true occupancy of 
invisible objects such as chairs and glass showcases. That work's scope was 
however limited to the exact setting where it was proposed. Only recently, mostly due 
to the growing possibilities of deep learning, a lot of new work has been 
proposed using neural networks trained on laser data 
\cite{sergeant_multimodal_2015,li_vehicle_2016,ondruska_deep_2016,engelcke_vote3deep_2017,liao_parse_2017,pfeiffer_perception_2017}.

In the context of obstacle detection, the use of neural networks has been 
proposed to infer bounding boxes of pedestrians and vehicles from 3D laser data 
of typical road scenes \cite{li_vehicle_2016,engelcke_vote3deep_2017} and to 
track spatially occluded moving objects in the scene by incorporating 
consecutive laser measurements to include the temporal information 
\cite{ondruska_deep_2016}. These works, however, are not directly dealing with 
the problem of inferring \truedist{}s. Moreover, 
\cite{li_vehicle_2016,engelcke_vote3deep_2017} use 3D laser sensors, which are 
naturally able to obtain the \truedist{}.

In regards to network architectures proposed for laser data processing, 
Pfeiffer \etal{} \cite{pfeiffer_perception_2017} trained a \ac{dnn} end-to-end 
to mimic a motion planner with laser data as input and velocity commands as 
output. The training data was obtained from simulation but validated on a real 
robot navigating complex corridors. In that work the authors propose a neural 
network 
architecture specifically tailored to process laser data: it consisted of a 
fully \ac{cnn}, using \acp{bn}, \acp{relu} and skip connections (details about 
these structures is given in \secref{sec:method}). Despite good results which 
validated the network structure, the authors claimed the learned motion planner 
underperformed when in wide open areas with a lot of glass and/or clutter, a 
problem that is partly due to missing obstacles in laser data and that we 
address in this work.

The problem of undetected obstacles when using 2D laser scans was also 
mentioned in the work by Liao \etal{} \cite{liao_parse_2017}, where they used 
2D laser scans together with RGB images to predict a 3D depth map 
of the same scene by a \ac{resnet}. 
Although the method was not per se developed for obstacle detection and 
avoidance, it was experimentally shown to successfully detect obstacles that 
were not seen from the fixed 2D laser. However, to produce such measurements 
the algorithm still relies on integrating information from RGB images.

Along the same line of inferring missing obstacles in laser data with vision, 
Baltzakis \etal{} \cite{baltzakis_fusion_2003} proposed a method which at 
runtime fuses laser and visual data, and is an example of sensor fusion. The 
method constructs a local 3D model from laser scans which, in turn, is visually 
evaluated by a stereo vision system to detect incorrect models. In such cases, 
the model is corrected with the depth data from the visual sensor. The method 
was successfully tested in simulation and on a real robot navigating a corridor 
filled with tables, chairs, and open doors. Although the problems addressed in 
sensor fusion works such as 
\cite{baltzakis_fusion_2003,liao_parse_2017} are in essence the same as in this 
work, those methods require overlapping readings from camera and laser sensors, 
which our method does not.

\section{Method}
\label{sec:method}

Let us define the output of a generic $N$-point 2D laser positioned at a height 
$h$ from floor level as a 1D vector $\matr{l}_h = \{l_{ih}\}_{i=1}^{N}$ where 
each $l_{ih}$ represents an estimate (usually, in meters) of the distance 
$d_{ih}$ of 
closest obstacle from the laser along the direction $i$ at height $h$.
When considering a specific robot model having a 2D laser sensor positioned at 
a fixed height $h^* \in [0,H]$, where $0$ is floor level and $H$ is the robot 
height, we define the vector $\matr{x} = \{x_{i} \mid 
x_i = l_{ih^*}\}_{i=1}^{N}$.

In this work we address the problem of predicting \truedist{}s, given an 
$N$-point 2D range laser acquisition $\matr{x}$. The \truedist{}s are 
represented as 
$\matr{y} = 
\{y_i\}_{i=1}^{N}$, \ie{} in the same space as the input laser $\matr{x}$. 
Formally, we can define each $y_i$ as
\begin{equation} \label{eq:yi}
	y_i = \min_{h \in [0,H]} d_{ih}\enspace.
\end{equation}

In particular, we want to learn the function $\mathcal{H} : \mathbb{R}^N 
\to 
\mathbb{R}^N $ for which $\mathcal{H}(\matr{x}) = \matr{y}$. To this end, 
following the intuition behind the work by Pfeiffer \etal{}
\cite{pfeiffer_perception_2017}, we propose to train a fully convolutional 
autoencoder architecture. An autoencoder is a deep learning architecture that 
can be formalized as a pair of transformations $\phi$ and $\psi$ such that:
\begin{equation} \label{eq:phipsi}
\begin{array}{c}
	\phi: \mathbb{R}^N \to \mathbb{R}^m\enspace,\\
	\psi: \mathbb{R}^m \to \mathbb{R}^N\enspace.
\end{array}
\end{equation}
Traditionally autoencoders were proposed in unsupervised 
settings \cite{bengio_learning_2009,testa_lightweight_2015} to reconstruct the input, \ie{} to compute the \emph{latent 
variable} $\matr{z} = \phi(\matr{x}) \in \mathbb{R}^m$ and then reconstructing 
the original input as $\matr{x} = \psi(\matr{z})$, given a loss function 
$\mathcal{L}$,
\begin{equation} \label{eq:lossae}
\phi, \psi = \underset{\phi,\psi}{\operatorname{arg\,min}}~\mathcal{L}((\psi 
\circ \phi)\matr{x}, \matr{x})\enspace.
\end{equation}
In that setting typically $m < N$, and the signal $\matr{x}$ is effectively 
compressed in the representation $\matr{z}$. For this reason, $\phi$ and $\psi$ 
are typically referred to as \emph{encoder} and \emph{decoder} respectively.

In contrast, we propose to use an autoencoder to predict $\matr{y}$ instead of reconstructing $\matr{x}$. This moves us from an unsupervised to a 
supervised learning setup, and for this reason knowledge of the correct 
$\matr{y}$, \ie{} a ground truth, is required to train the network. We will 
discuss in detail how we propose to empirically obtain $\matr{y}$ in 
\secref{sec:dataset}.

\subsection{Network architecture}
\label{sec:net}

\begin{figure*}
	\centering
	\hspace{-1.9em}	\tikzstyle{layer} = [fill=purple!20,draw=black]
	\tikzstyle{conv} = [thick,->,>=stealth,double]
	\tikzstyle{func} = [thick,->,>=stealth]
	\tikzstyle{skip} = [thick,dashed,>=stealth,->]
	\tikzstyle{plus} = [draw,circle,fill=white,inner sep=.5pt,scale=.55]
	
	\begin{tikzpicture}[scale=0.41]
		\scriptsize
		\filldraw[layer] (0,0) rectangle (1,4);
		\filldraw[layer] (2,1) rectangle (4,3);
		\filldraw[layer] (5,1.5) rectangle (9,2.5);
		\filldraw[layer] (10,1.75) rectangle (18,2.25);
		\filldraw[layer] (20,1.75) rectangle (28,2.25);
		\filldraw[layer] (29,1.5) rectangle (33,2.5);
		\filldraw[layer] (34,1) rectangle (36,3);
		\filldraw[layer] (37,0) rectangle (38,4);
		
		\node (x) at (-2,2) {$\matr{x}/s$};
		\node (z) at (19,2) {$\matr{z}$};
		\node (y) at (40,2) {$\matr{\hat{y}}/s$};
		
		\draw[conv] (x) -- (0,2);
		\draw[conv] (1,2) -- (2,2);
		\draw[conv] (4,2) -- (5,2);
		\draw[conv] (9,2) -- (10,2);
		\draw[conv] (28,2) -- (29,2);
		\draw[conv] (33,2) -- (34,2);
		\draw[conv] (36,2) -- (37,2);
		\draw[conv] (38,2) -- (y) node[midway,anchor=south west] {$\gamma$};
		
		\node at (0.5,-1) {$64\times16$};
		\node at (3,0) {$32\times32$};
		\node at (7,0.5) {$16\times64$};
		\node at (14,0.75) {$8\times128$};
		\node at (24,0.75) {$8\times128$};
		\node at (31,0.5) {$16\times64$};
		\node at (35,0) {$32\times32$};
		\node at (37.5,-1) {$64\times16$};
		
		\draw[skip] (1.5,4) to[out=10,in=170] (36.5,4);
		\draw[skip] (4.5,3) to[out=10,in=170] (33.5,3);
		\draw[skip] (9.5,2.5) to[out=10,in=170] (28.5,2.5);
		
		\draw[decorate,decoration={brace}] (18,-2) -- (-1,-2)
			node[midway,yshift=-1.5em] {$\phi$};
		\draw[decorate,decoration={brace}] (39,-2) -- (20,-2)
			node[midway,yshift=-1.5em] {$\psi$};
		\node[plus] at (19,3.5) {$+$};
		\node[plus] at (19,4.5) {$+$};
		\node[plus] at (19,5.8) {$+$};
			
		\node at (0, -4.5) {Legend:};
		\draw[conv] (2,-4.5) -- (3,-4.5) node[anchor=west,xshift=.5em] 
			{convolution (kernel size: 5, sdride: 2) + batch normalization + 
			ReLU};
		\draw[skip] (21.5,-4.5) -- (24,-4.5) node[anchor=west,xshift=.5em] 
			{skip connection};
		\node[plus] at (22.75,-4.5) {$+$};
	\end{tikzpicture}
	\caption{\label{fig:nn}The proposed fully convolutional autoencoder}
\end{figure*}
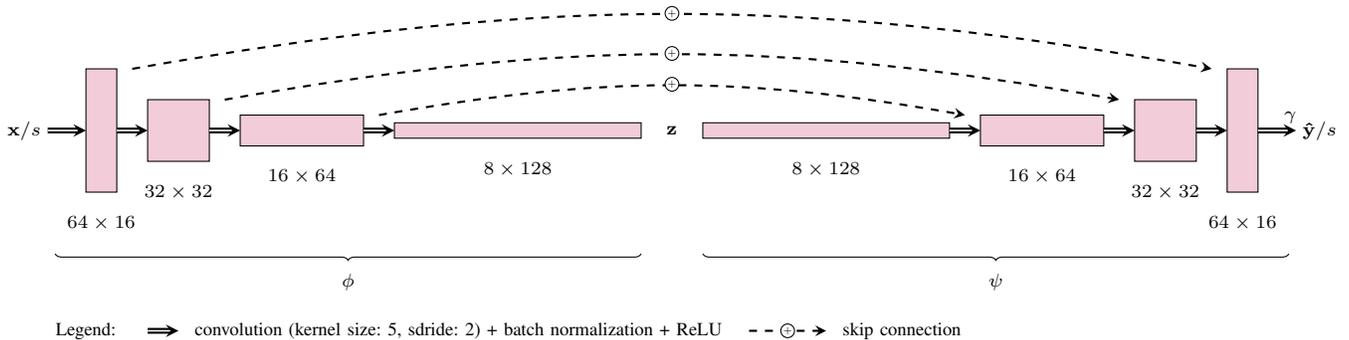

A graphical representation of the proposed network structure is given in 
\figref{fig:nn}. The encoder $\phi$ takes as input a 1D vector $\matr{x}$ of 
size $N = 128$ and passes it through four 1D convolutional layers, with kernel 
size 5 and stride 2, connected via \ac{bn} \cite{ioffe_batch_2015} and 
\ac{relu} \cite{he_delving_2015} layers. The input vector $\matr{x}$ is scaled 
to the range $[0,1]$ before feeding it to the network. The output of the 
encoder is reshaped to obtain the latent feature representation $\matr{z}$. The 
decoder $\psi$ has the same structure as $\phi$, which maps $\matr{z}$ back to 
a 1D vector $\matr{\hat{y}}$ of size $N$ through a series of 1D transposed 
convolutional (often, improperly, referred to as deconvolutional) layers of 
kernel size 5 and stride 2 connected via \ac{bn} and \ac{relu}.
In the next sections we will discuss some components of the network 
architecture and the intuitions rooted in the specific laser data domain that 
justifies them. The actual contribution of these components to the performance 
of the network is evaluated in \secref{sec:experiments}.

\subsection{Skip connections}
\label{sec:skip}

We propose to connect corresponding convolutional and deconvolutional layers 
through skip connections, as shown in \figref{fig:nn}. Previous works 
\cite{mao_image_2016,newell_stacked_2016,pfeiffer_perception_2017} demonstrated 
a two-folded contribution of such connections: firstly they improve gradient 
back-propagation to bottom layers and secondly they increase the flow of detail 
information to top layers.

The way skip connections work is that the output of a convolutional 
layer in the encoder is added to the output of the corresponding 
deconvolutional layer of the same size in the decoder. The result is then used 
as the input of the next deconvolutional layer. This operation allows high 
frequencies that are typically lost by the encoder to be passed to the decoder 
to produce a more detailed reconstruction.

\subsection{Non-uniform scaling}
\label{sec:gamma}

Both the input laser $\matr{x}$ and the output obstacle vector $\matr{y}$ 
represent distance measures in meters with values in range $[0,s]$ where 
$s$ is the maximum laser distance (in our work, 30 m). Intuitively, for the 
robot to safely navigate, an error in the distance estimate for a closer 
object is far more dangerous than for an object far away; for this reason 
accurate prediction in the first couple of meters is crucial. However, when 
considering the range of a typical laser, the first couple of meters are 
residing in a small fraction of the range. To address this issue, we 
propose to first scale the input in the range $[0,1]$ by linear scaling, and then 
scale the output of the last layer of the network $\matr{\hat{y}'}$ by a 
positive $\gamma$ factor, before bringing it back to its original range. 
Formally,
\begin{eqnarray} \label{eq:gamma}
\matr{x'} &=& \matr{x}/s\enspace,\\
\matr{\hat{y}'} &=& (\psi \circ \phi)\matr{x'}\enspace,\\
\matr{\hat{y}} &=& s(\matr{\hat{y}'})^\gamma\enspace.
\end{eqnarray}

The intuition behind this non-uniform output scaling, which henceforth is referred to as 
$\gamma$-scaling, comes from the domain of image processing, where \emph{gamma 
correction} is used to enhance contrast in an image and \emph{gamma encoding} 
is used to optimize the bandwidth used to transport an image according to the 
way humans perceive light and color.

\begin{figure}
	\centering
		\begin{tikzpicture}
		\scriptsize
		\begin{axis}[
			height=6cm,
			width=6cm,
			axis x line=bottom,
			axis y line=left,
			xlabel=$\matr{\hat{y}'}$,
			ylabel=$\matr{\hat{y}}/s$,
			ymin=0,ymax=1,
			xmin=0,xmax=1,
			legend pos=outer north east,
			legend style={draw=none},
		]
		\addplot[domain=0:1,samples=200,color=purple] {x^.25};
		\addplot[domain=0:1,samples=200,color=red] {x^.5};
		\addplot[domain=0:1,color=black] {x};
		\addplot[domain=0:1,samples=200,color=blue] {x^2};
		\addplot[domain=0:1,samples=200,color=green] {x^4};
		\legend{$\gamma = 1/4$,$\gamma = 1/2$,$\gamma = 1$,$\gamma = 
			2$,$\gamma = 4$}
		\end{axis}
	\end{tikzpicture}
	\caption{\label{fig:gamma}Effect of different values of $\gamma$ on the 
		output of the last layer of the network.}
\end{figure}
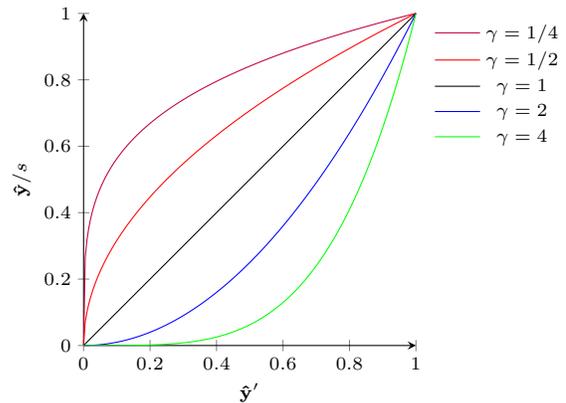

The effect of different values of $\gamma$ is shown in 
\figref{fig:gamma}. By using $\gamma > 1$, a greater portion of the range of 
the last layer of the network is reserved for the smallest values of the 
output, effectively increasing the resolution in that portion of the range 
while reducing the resolution for the rest of the output range. Using $0 < 
\gamma < 1$ yields the opposite effect instead, which in our case is not 
desired.

\begin{figure*}
	\centering
	\includeexample{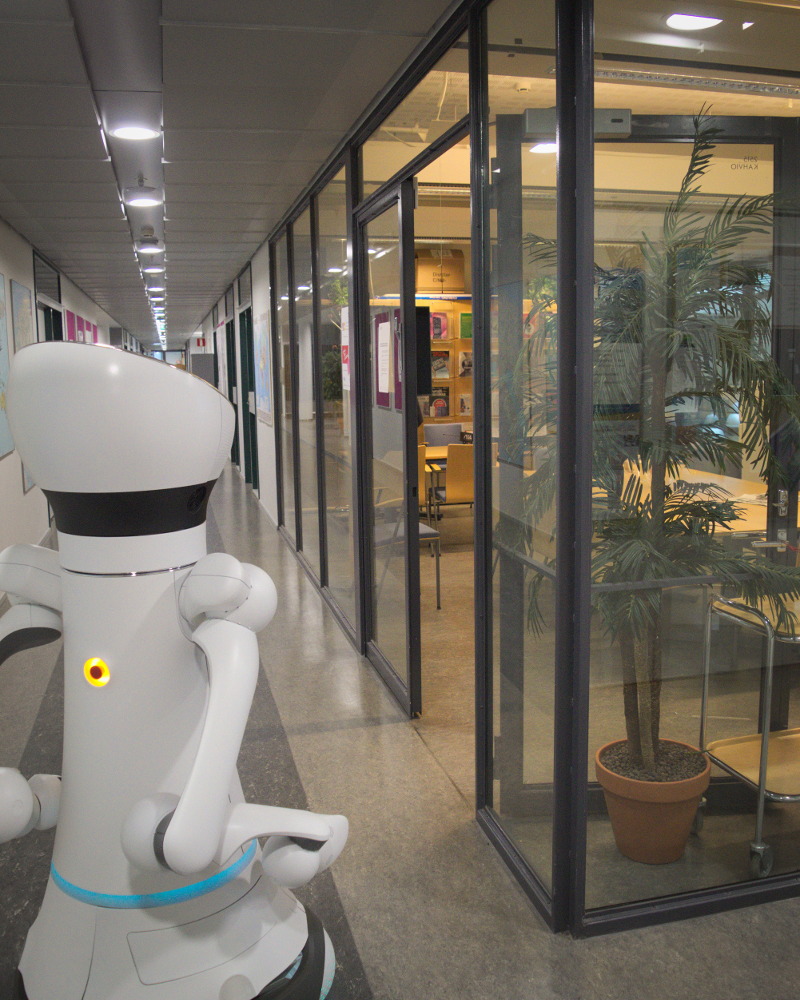}{Room with glass walls}%
	\hspace{-2.1em}\includeexample{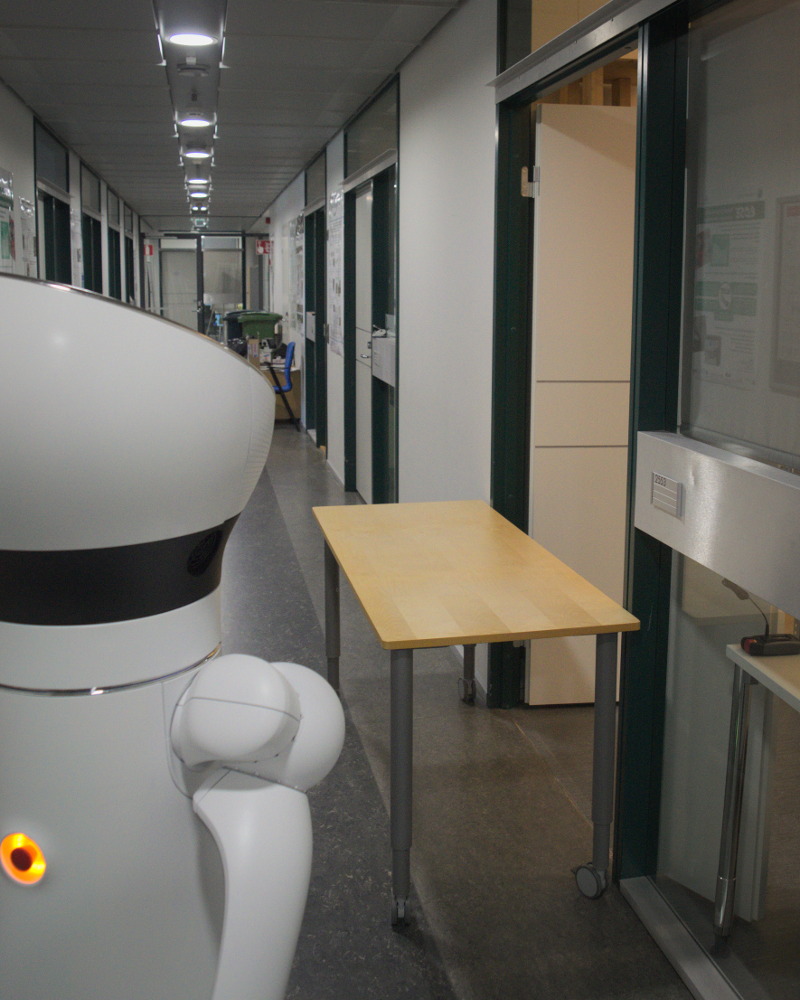}{Table blocking a door}%
	\hspace{-2.1em}\includeexample{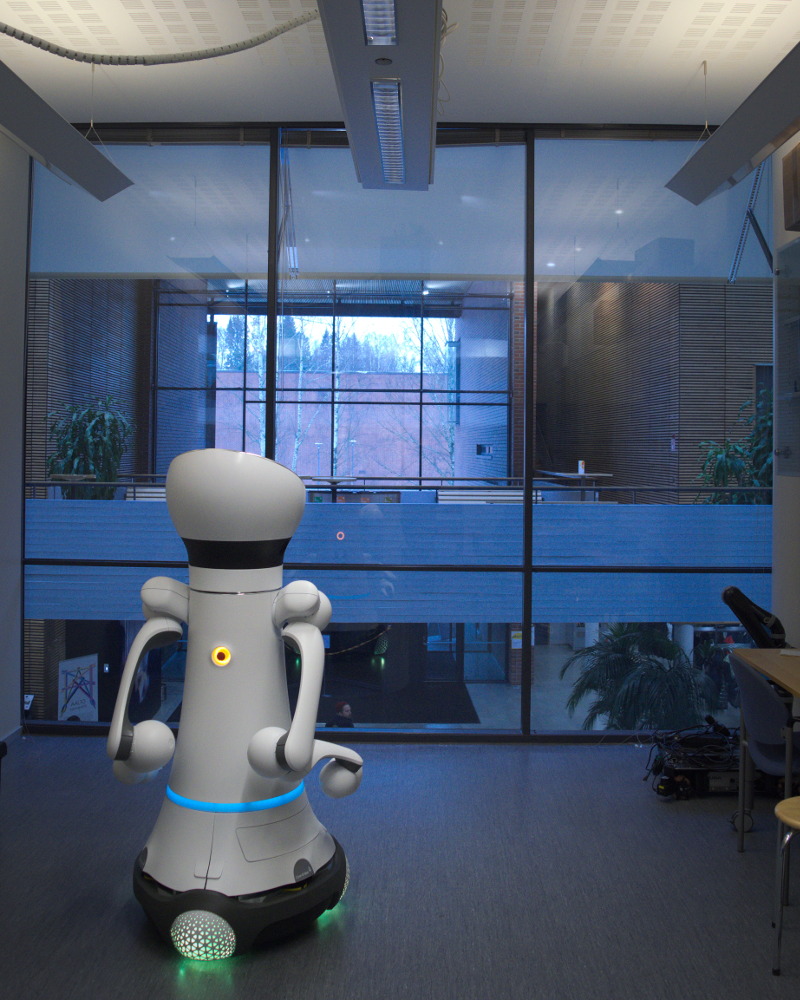}{Glass window}%
	\hspace{-2.1em}\includeexample{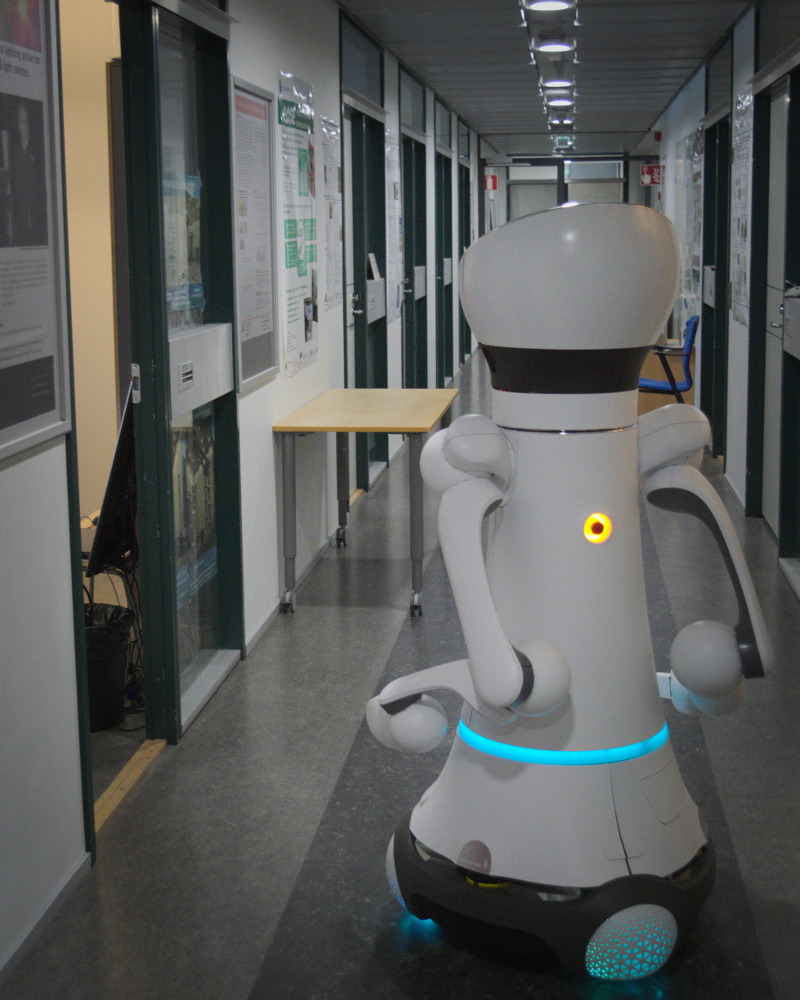}{Table in a corridor}%
	\hspace{-2.1em}\includeexample{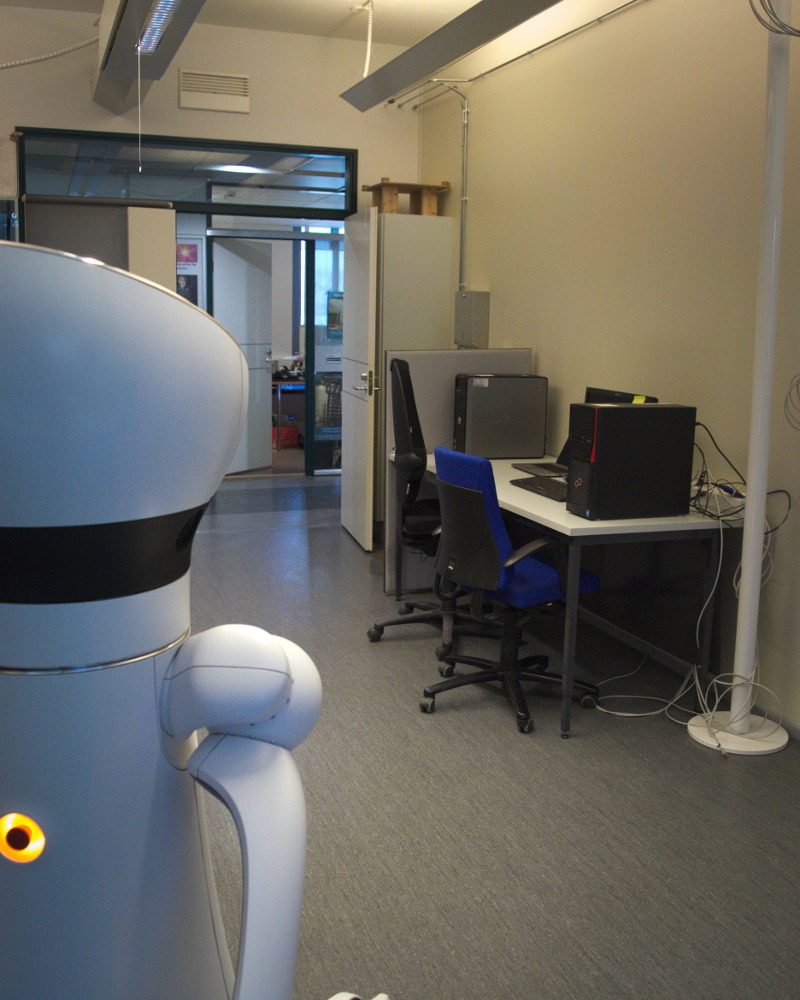}{Table and chairs}%
	\caption{\label{fig:examples}Examples of scenes encountered while testing 
		on the \cob{}; \laserlegend{} \bestcolor{}.}
\end{figure*}

\subsection{Loss function}
\label{sec:loss}

Following the same reasoning as in \secref{sec:gamma}, we trained the 
network using \ac{rmsle} as loss function.
\ac{rmsle} can be expressed as:
\begin{equation} \label{eq:rmsle}
	\begin{array}{rcl}
		\mathcal{L}(\matr{\hat{y}}, \matr{y}) & = & 
		\sqrt{\frac{1}{N}\sum_{i=1}^{N}[\ln(\hat{y}_i+1) - 
			\ln(y_i+1)]^2}\\[.5em]
		& = & 
		\sqrt{\frac{1}{N}\sum_{i=1}^{N}\ln^2\frac{\hat{y}_i+1}{y_i+1}}\enspace;
	\end{array}
\end{equation}
it is a common choice for domains in which the desire is to mitigate the 
contribution of big errors when numbers are big themselves.

It is important to note that by using $\gamma > 1$, the network imposes variable output resolution to produce smaller errors for smaller numbers while admitting larger 
errors for bigger numbers. It is 
therefore important to pair this choice with a loss function that follows the 
same idea. If \ac{mse} was used, the error for big numbers 
could possibly overshadow more important smaller errors and negatively impact 
the learning process.

\section{Experiments and results}
\label{sec:experiments}

\subsection{Experimental Platform}
\label{sec:platform}

In this work we used the mobile service robot \cob{} shown in \figref{fig:cob} 
and \figref{fig:examples}. The robot is equipped with omni-directional wheels 
and three 2D SICK S300 safety laser scanners, one at front, one to the left, 
and one to the right, with maximum scanning range of 30 m, resolution of 30 mm 
and angular resolution of \ang{0.5}. Together all three lasers scanners enable 
the robot to sense its \ang{360} surrounding. Furthermore, the robot is 
equipped with three RGBD Asus Xtion cameras located at the front of the robot 
together covering a field of view of approximately \ang{90}. As a consequence, 
the robot senses only about \ang{90} of its surrounding with overlapping laser 
and image sensors while the remaining \ang{270} are sensed using laser scanners 
alone. 
This sensing limitation motivated the work in this paper as our method enables 
inferring \truedist{}s from laser data without requiring additional sensors. 

\subsection{Dataset}
\label{sec:dataset}

\begin{figure}
	\centering
		\begin{tikzpicture}
		\scriptsize
		\begin{axis}[
			width=7cm, 
			height=6cm,
			axis x line=bottom,
			axis y line=left,
			xtick distance=20,ytick distance=5,
			xmin=0, xmax=125,
			ymin=0, ymax=30,
			xlabel=$i$,
			ylabel=distance (m),
			legend pos=north east,
			legend style={draw=none},
		]
		\addplot[thick,green] table[x expr=\coordindex, 
		y=x]{Figures/sample1.csv};
		\addplot[ultra thick,dashed,blue] table[x expr=\coordindex, 
		y=yd]{Figures/sample1.csv};
		\addplot[thick,red] table[x expr=\coordindex, y=y]{Figures/sample1.csv};
		\legend{$\matr{x}$,$\matr{y}^c$,$\matr{y}$}
		\end{axis}
	\end{tikzpicture}
	\caption{\label{fig:datasamples}Sample data from the training dataset. This 
		reading is taken while the robot is facing a closed two-panel glass 
		door.}
\end{figure}
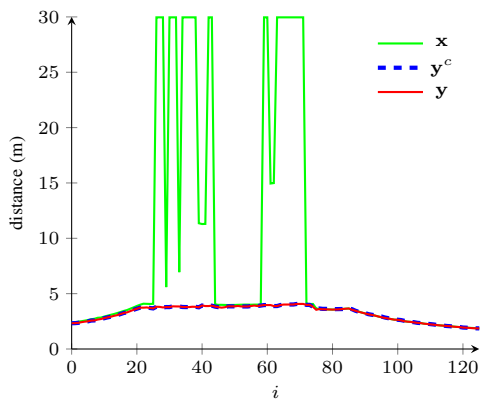

As already mentioned in \secref{sec:method}, we are training in a supervised 
learning setting, which requires matching pairs of input laser 
data $\matr{x}$ and \truedist{} vectors $\matr{y}$. We therefore 
acquired a dataset composed of laser scans, serving as input, and corresponding 
depth images of the same scene obtained from one of the front-facing cameras on 
the \cob{}. Then, the depth images were projected to 1D by iteratively taking 
the closest distance of each column in the depth image and storing it as the 
distance vector estimated by the depth camera. Furthermore, to focus on obstacles that the robot could collide with, we removed from the 
depth images all points that, once converted to the 3D space, were outside the range $[\epsilon, H]$ with $\epsilon = 0.05 \unit{m}$. 
Additionally, as the field of view of a single laser scanner is about \ang{270} 
while only \ang{90} 
are covered by the RGBD sensors, it was necessary to prune the laser scan until 
it matches the field of view of the camera. Once the matching was complete we 
obtained two 1D vectors of size $N=128$, one originated from the laser scanner 
and one obtained from the depth camera, which we will refer as $\matr{x}$ and 
$\matr{y}^c$ respectively. Then we set $\matr{y}$ as
\begin{equation}\label{eq:y}
\matr{y} = \{y_i = \min(x_i, y^c_i)\}_{i=1}^N\enspace.
\end{equation}
The final step forces the ground truth to consist of the closest points from either sensor readings to provide a conservative estimate of the obstacle distance. An example of $\matr{x}$, $\matr{y}^c$, and $\matr{y}$ is shown in \figref{fig:datasamples}.

This approach of generating the ground truth from depth images provided us with 
a good estimate of the \truedist{}, did not require manual labeling, and 
enabled us to acquire a dataset directly by running the robot around. However, 
generating similar ground truth would be also possible from, for example, an 
accurate map of the environment, or human labeling.

We gathered the dataset by teleoperating the \cob{} in a university building. 
We gathered a training set consisting of 27904 input-output pairs in the second 
floor of the building, while the test set was obtained in the third floor and 
consists of 11604 input-output pairs. Several of these locations included glass 
surfaces, tables and chairs, \eg{} those shown in \figref{fig:examples}, 
whose \truedist{} is incorrectly inferred by the laser scanner but 
correctly estimated by the depth image, capturing interesting evaluating 
scenarios for the network to represent. Although the two floors present structural 
similarities, we argue they vary enough in the overall topology and obstacle 
content to act as a good train-test set pair. 
Nevertheless, in future works we aim at expanding both the training and test 
set with data from various indoor environments.

\subsection{Data augmentation}
\label{sec:augmentation}

In other domains \cite{holmstrom_using_1992,matsuoka_noise_1992,yin_noisy_2015} it is known that injection of noise in 
the input of the training dataset can improve the generalization capability of 
neural networks and encourages a network to converge to smoother mapping 
functions. However, to the best of our knowledge, this has never been tested 
when training on laser. Therefore, during training, we augmented the data by 
adding random noise sampled from a Gaussian distribution $\mathcal{N}(0, 
\sigma_n)$ to the input. In this study, we set $\sigma_n = 0.02 \unit{m}$, 
which is compatible with the sensor resolution of the \cob{}. In the next 
section we evaluate and discuss the contribution of noise injection on training 
performance.

In addition, to further increase variability in the training set, for each 
training example, a random 50\% chance of flipping both $\matr{x}$ and 
$\matr{y}$ has been implemented. This way, the number of samples in the 
training dataset effectively doubles.

\begin{figure}
	\centering
	\includegraphics[width=.9\linewidth]{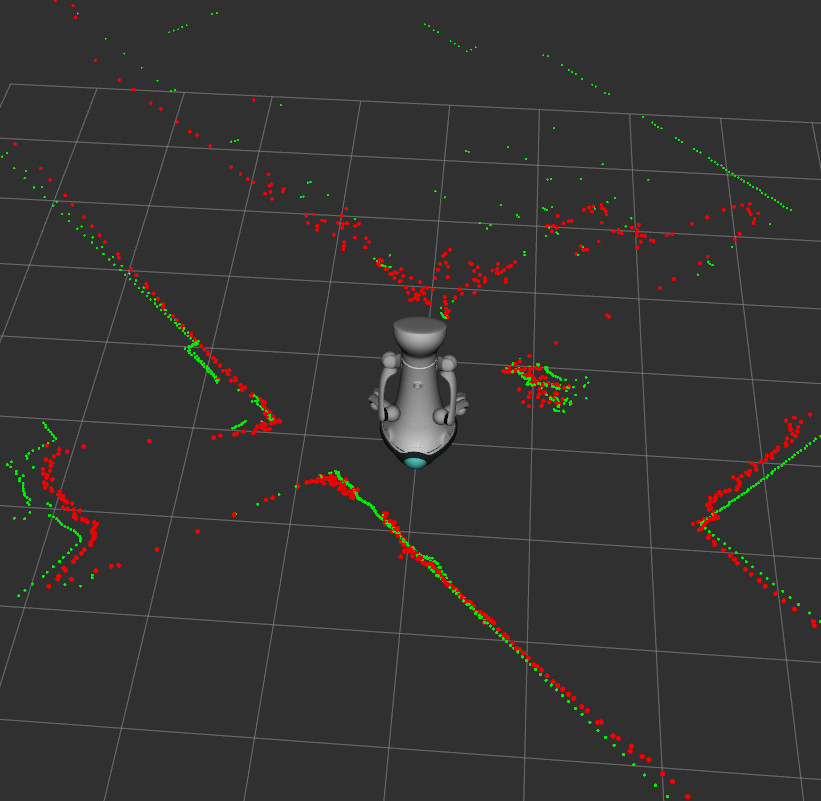}
	\caption{\label{fig:360}\ang{360} laser hallucination; \laserlegend{} 
		\bestcolor{}.}
\end{figure}

\subsection{Evaluation on test set}
\label{sec:quantitative_eval}

All networks evaluated in this work were implemented in pyTorch 0.3.0 and 
trained for 2000 epochs on the augmented training dataset presented in the 
previous sections by using shuffled batches of 32 samples. The networks were 
trained using \ac{rmsle} as loss function and optimizing with \ac{adam} 
\cite{kingma_adam_2014}.

\begin{table*}
	\centering
	\caption{\label{tab:results}Test-set performance and parameter 	
		contributions}
	
	\begin{tabular}{cccccccc}
		\toprule
		& \multicolumn{3}{c}{Network hyperparameters} & 
		\multicolumn{2}{c}{\ac{rmsle} (absolute)} & 
		\multicolumn{2}{c}{\ac{rmsle} (relative to n. 0)} \\
		n. & Skip connections & $\gamma$ & $\sigma_n$ & mean ($\times 10^{-2}$) 
		& 
		std ($\times 10^{-3}$) & mean & 
		std \\
		\midrule
		0 & \true{} & 2 & 0.02 & 2.865 & 0.31 & & \\
		\midrule
		1 & \false{} & 2 & 0.02 & 3.059 & 1.45 & +6.79\% & +372.81\% \\
		2 & \true{} & 2 & \false{} & 2.838 & 0.57 & -0.95\% & +84.82\% \\
		3 & \true{} & \textonehalf & 0.02 & 2.885 & 1.27 & +0.71\% & 
		+315.13\% 
		\\
		4 & \true{} & 1 & 0.02 & 2.938 & 1.01 & +2.54\% & +230.12\% \\
		5 & \true{} & 4 & 0.02 & 2.869 & 0.47 & +0.13\% & +52.35\% \\
		6 & \false{} & 1 & \false{} & 3.065 & 0.76 & +6.98\% & +148.91\% \\
		\bottomrule
	\end{tabular}
\end{table*}

We are in particular interested in evaluating whether skip connections, 
$\gamma$-scaling and the injection of noise increase the performance. To 
quantitatively evaluate their contributions and the effective 
ability of the proposed architecture to learn the true obstacle distances from 
laser data, we trained each network configuration 5 times. The results, in 
terms of mean and standard deviation of the \ac{rmsle}, are presented in 
\tabref{tab:results}. The configuration n. 0 is taken as baseline, \ie{} the 
one including all hyperparameters presented in this work. Configurations n. 1-5 
are obtained by changing only one of the hyperparameters at the time, while 
configuration n. 6 is obtained by having all proposed hyperparameters disabled. 
This approach enabled us to evaluate the contribution of each single 
component to the performance.

From the results presented in \tabref{tab:results}, it is clearly seen that 
among all hyperparameters, the inclusion of skip connections improves the 
network performance the most. Additionally, it can be concluded that using a 
value of $\gamma > 1$ is a good choice for this domain. However, as is 
predictable, a 
too high value, as in the case of $\gamma = 4$, limits improvements due to 
exaggerated scaling. Finally, it is 
interesting to notice the effect of the inclusion of noise: the negligible 
deficit to the average performance it produces is counterbalanced by an almost 
halved standard deviation; which seems to confirm the results presented in 
\cite{holmstrom_using_1992,matsuoka_noise_1992,yin_noisy_2015}, where the main 
contribution of injecting noise into the training inputs was claimed to be an 
increased ability of the network to learn more consistent results.
Configuration n. 6, \ie{} the one having all hyperparameters excluded, performs 
the worst, confirming once more the positive effect all proposed 
hyperparameters have on performance.

\subsection{Online testing on the \cob{}}
\label{sec:qualitative_eval}

Based on the quantitative results in \tabref{tab:results}, the network with 
skip connections, $\gamma=2$, and trained with zero mean Gaussian noise with 
standard deviation $\sigma_n = 0.02$ provided the best results. The same 
network was further tested online on \cob{} in the same office spaces from 
which the training and test sets were gathered. The reason for 
testing on the same floor as the training set was because this floor 
included many challenging situations such as the room with glass walls and 
doors presented in \figref{fig:examples_coffee}. Additionally we changed the environment by, for example, relocating tables to close access to 
corridors (\figref{fig:cob}) and doors (\figref{fig:examples_door}), creating situations not present in the training set. Other test scenarios included a glass window covering an entire wall
(\figref{fig:examples_window}) and a table with chairs underneath 
(\figref{fig:examples_office}).

The figures show that the trained network is able to hallucinate laser 
data that better estimates the \truedist{}. Particularly impressive results 
occur when the robot faces the room with glass walls in \figref{fig:examples_coffee} and 
correctly recognizes the door from other walls. It is also worth pointing out 
that the network can infer missing laser data from a small set of points 
corresponding to obstacles such as the table in \figref{fig:examples_table}. 
Although the network was trained on data gathered from a \ang{90} viewing 
angle, it is possible to hallucinate \ang{360} around the robot as shown in 
\figref{fig:360} by running the network on laser chunks of that size.

\begin{figure}
	\includegraphics[width=.49\linewidth]{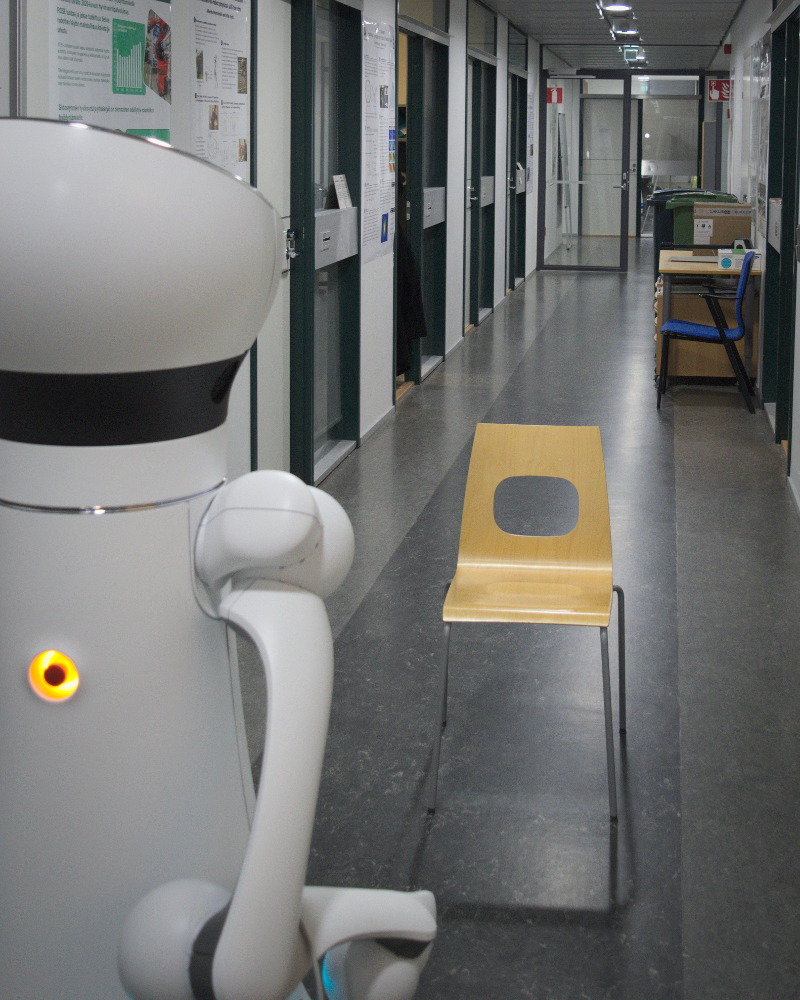}
	\includegraphics[width=.49\linewidth]{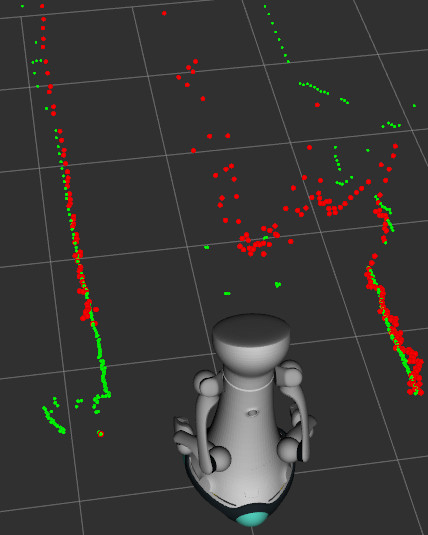}
	\caption{\label{fig:chair}A problematic sample; \laserlegend{} 
		\bestcolor{}.}
\end{figure}

In most testing situations the proposed approach produced correct results, even when we changed object locations to ones 
the network never saw during training. Nevertheless, we noticed that the trained network was unable to correctly infer \truedist{} in scenarios greatly different 
from the training set. For example, tables were correctly detected even if 
moved to new locations as long as they were parallel to a wall, but if placed 
perpendicular they were not correctly detected.
Similarly, as shown in 
\figref{fig:chair}, objects positioned alone in the middle of empty spaces were 
not correctly detected. Both these situations were not present in the training 
set and were therefore not learned by the network. To address this issue we 
want to explore more thorough data augmentation in the future, for example by 
synthetically injecting objects in random positions in the data. Another interesting network behavior is the tendency to produce smooth curves 
even in presence of strong discontinuities. This phenomenon is visible for the doors shown in \figref{fig:360}, the corridor however was correctly kept open. Although such behavior is not 
problematic most of the times, it can cause navigation issues when planning through a door, as some navigation goals may seem unreachable. The smoothing tendency stems from the network topology, and to overcome it one solution is to more aggressively enforce non-linearities in the network structure.

\begin{figure}
	\centering
	\subfloat[]{\includegraphics[width=.32\linewidth]{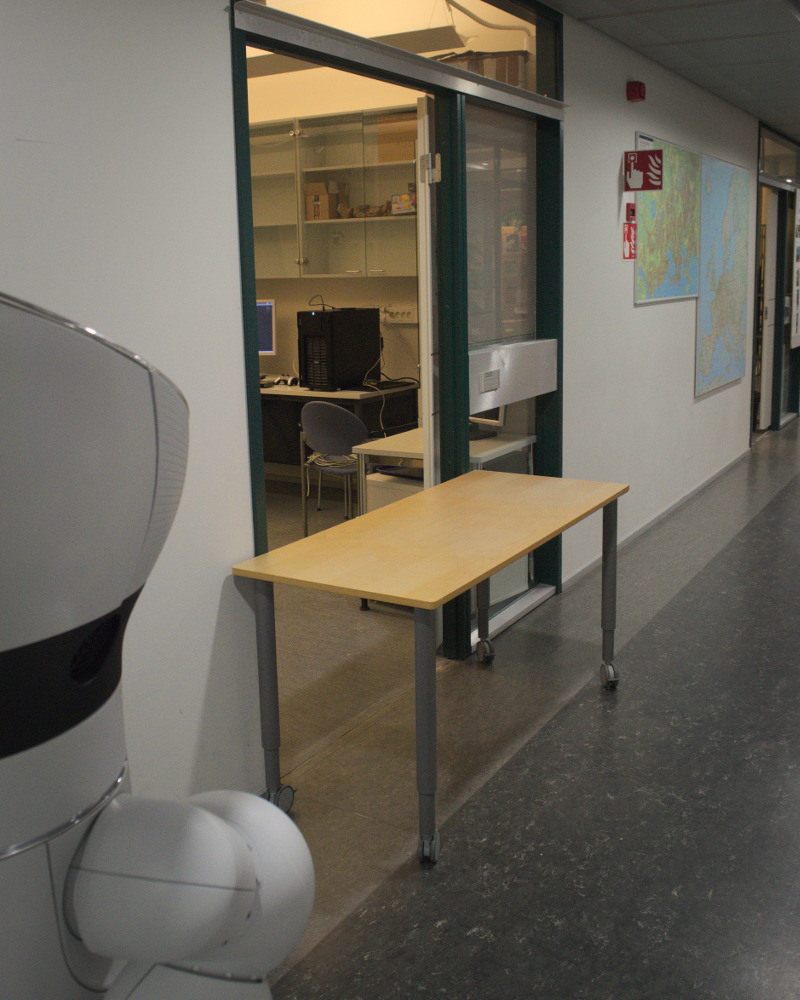}\label{fig:plan_photo}}~
	\subfloat[]{\includegraphics[width=.32\linewidth]{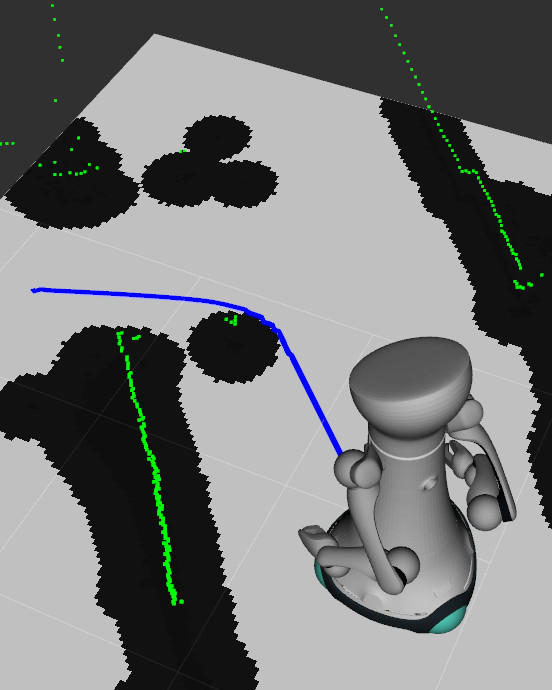}\label{fig:plan_laser}}~
	\subfloat[]{\includegraphics[width=.32\linewidth]{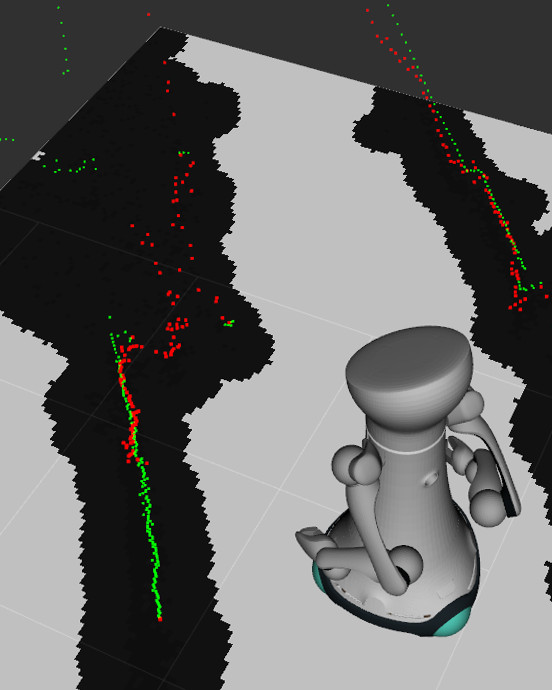}\label{fig:plan_hall}}
	\caption{\label{fig:planner}Comparison of the behaviour of a local planner 
		relying on raw laser data (b) or on hallucinated laser data (c) while 
		the 
		robot is instructed to plan a trajectory through a door obstructed by a 
		table; in both pictures the white area is considered free space, the 
		black 
		areas are considered obstructed, the blue line is the planner 
		trajectory, 
		\laserlegend{} \bestcolor{}.}
\end{figure}

Finally, we tested the use of the hallucinated laser for autonomous local 
navigation. \figref{fig:planner} shows an example where the robot was given a 
navigation goal inside the room. The planner using only raw laser 
planned a trajectory through the table (the blue line in 
\figref{fig:plan_laser}) which would cause a collision, while the one relying on the hallucinated laser did not as it successfully 
inferred the actual \truedist{} as shown in \figref{fig:plan_hall}.

\addtolength{\textheight}{-2.5cm}
\section{Conclusions and future work}
\label{sec:conclusions}

Detecting and inferring distance to complex obstacles from 2D laser scans 
alone is nontrivial but useful for robot safety. In this work, we presented a 
method capable of estimating the \truedist{} from raw 2D laser data. 
This is achieved by framing the problem as a learning task and training a 
\ac{dnn} to infer the  \truedist{}s from the laser scans. The 
key concept is to shape the \ac{dnn} as an autoencoder with skip connections 
between each convolutional and deconvolutional layer, enabling the network to 
pass detailed information from bottom to top layers. Furthermore, to generate 
the actual training and test set we proposed a method to estimate the true 
distance to obstacles by fusing 2D laser data and depth images. Based on the 
quantitative evaluation on the test set, the network with skip connections 
outperformed the ones without, and by further adding noise to the input data 
and shaping the last layer using a novel non-uniform $\gamma$-scaling the 
overall performance increased while the standard deviation decreased, resulting 
in more robust learning. The real robot test demonstrated that the trained 
network was able to infer the \truedist{} in \ang{360} in challenging 
situations including an entire wall made of glass windows or from a scarce set 
of laser points that corresponded to the obstacle as in the case of tables.

To the best of our knowledge no prior work has focused on problems similar to 
the one presented here, opening up interesting future research avenues. To 
improve generalization, the training dataset could be augmented by domain 
randomization, corresponding to changing location and 
rotation of objects such as tables, chairs, or windows. To enable prediction of 
moving obstacles, for example walking humans, a temporal component could be 
included in the network along the lines of 
\cite{ondruska_deep_2016}. Finally, the ideas presented in this work could be 
integrated with  end-to-end training for navigation, by extending a motion 
planner similar to the one in \cite{pfeiffer_perception_2017}.



\newpage

\bibliographystyle{IEEEtran}
\bibliography{biblio}

\end{document}